\documentclass[sigconf]{acmart}
\AtBeginDocument{%
  \providecommand\BibTeX{{%
    \normalfont B\kern-0.5em{\scshape i\kern-0.25em b}\kern-0.8em\TeX}}}

\setcopyright{acmlicensed}
\copyrightyear{2024}
\acmYear{2024}
\acmDOI{10.1145/3664647.3686836}

\acmConference[MM '24]{Proceedings of the 32nd ACM International Conference on Multimedia}{October 28-November 1, 2024}{Melbourne, VIC, Australia}
\acmBooktitle{Proceedings of the 32nd ACM International Conference on Multimedia (MM '24), October 28-November 1, 2024, Melbourne, VIC, Australia}
\acmISBN{979-8-4007-0686-8/24/10}

\usepackage{bm}
\usepackage{pgfplots}
\usepackage{subcaption}
\usepackage{tikz}
\usepackage{multirow}
\usepackage{pifont}

\pgfplotsset{compat=newest}

\definecolor{mygray}{gray}{.9}
\definecolor{sota_blue}{HTML}{0071bc}
\makeatletter
\newcommand{\thickhline}{%
	\noalign {\ifnum 0=`}\fi \hrule height 1pt
	\futurelet \reserved@a \@xhline
}
\makeatother

\begin{document}

\title{Dynamic Prompting of Frozen Text-to-Image Diffusion Models for Panoptic Narrative Grounding}


\author{Hongyu Li}
\authornote{Both authors contributed equally to the paper}
\affiliation{%
  \institution{School of Artificial Intelligence, Beihang University}
  % \streetaddress{1 Th{\o}rv{\"a}ld Circle}
  \city{Beijing}
  \country{China}}
\email{19377211@buaa.edu.cn}

\author{Tianrui Hui}
\authornotemark[1]
\affiliation{%
  \institution{School of Computer Science and Information Engineering, Hefei University of Technology}
  % \streetaddress{1 Th{\o}rv{\"a}ld Circle}
  \city{Hefei}
  \country{China}}
\email{huitianrui@gmail.com}

\author{Zihan Ding}
\affiliation{%
  \institution{School of Artificial Intelligence, Beihang University}
  % \streetaddress{1 Th{\o}rv{\"a}ld Circle}
  \city{Beijing}
  \country{China}}
\email{dingzihan737@gmail.com}

\author{Jing Zhang}
\affiliation{%
  \institution{School of Software, Beihang University}
  % \streetaddress{1 Th{\o}rv{\"a}ld Circle}
  \city{Beijing}
  \country{China}}
\email{zhang_jing@buaa.edu.cn}

\author{Bin Ma}
\affiliation{%
  \institution{Meituan}
  % \streetaddress{1 Th{\o}rv{\"a}ld Circle}
  \city{Beijing}
  \country{China}}
\email{mabin04@meituan.com}

\author{Xiaoming Wei}
\affiliation{%
  \institution{Meituan}
  % \streetaddress{1 Th{\o}rv{\"a}ld Circle}
  \city{Beijing}
  \country{China}}
\email{weixiaoming@meituan.com}

\author{Jizhong Han}
\affiliation{%
  \institution{Institute of Information Engineering, Chinese Academy of Sciences}
  % \streetaddress{1 Th{\o}rv{\"a}ld Circle}
  \city{Beijing}
  \country{China}}
\email{hanjizhong@iie.ac.cn}

\author{Si Liu}
\authornote{Corresponding author}
\affiliation{%
  \institution{School of Artificial Intelligence, Beihang University}
  % \streetaddress{1 Th{\o}rv{\"a}ld Circle}
  \city{Beijing}
  \country{China}}
\email{liusi@buaa.edu.cn}

\renewcommand{\shortauthors}{Hongyu Li and Tianrui Hui, et al.}

\begin{abstract}
  Panoptic narrative grounding (PNG), whose core target is fine-grained image-text alignment, requires a panoptic segmentation of referred objects given a narrative caption. Previous discriminative methods achieve only weak or coarse-grained alignment by panoptic segmentation pretraining or CLIP model adaptation. Given the recent progress of text-to-image Diffusion models, several works have shown their capability to achieve fine-grained image-text alignment through cross-attention maps and improved general segmentation performance. However, the direct use of phrase features as static prompts to apply frozen Diffusion models to the PNG task still suffers from a large task gap and insufficient vision-language interaction, yielding inferior performance. Therefore, we propose an Extractive-Injective Phrase Adapter (EIPA) bypass within the Diffusion UNet to dynamically update phrase prompts with image features and inject the multimodal cues back, which leverages the fine-grained image-text alignment capability of Diffusion models more sufficiently. In addition, we also design a Multi-Level Mutual Aggregation (MLMA) module to reciprocally fuse multi-level image and phrase features for segmentation refinement. Extensive experiments on the PNG benchmark show that our method achieves new state-of-the-art performance.
\end{abstract}

\begin{CCSXML}
<ccs2012>
   <concept>
       <concept_id>10010147.10010178.10010224.10010245.10010247</concept_id>
       <concept_desc>Computing methodologies~Image segmentation</concept_desc>
       <concept_significance>300</concept_significance>
       </concept>
   <concept>
       <concept_id>10010147.10010178.10010224.10010225.10010227</concept_id>
       <concept_desc>Computing methodologies~Scene understanding</concept_desc>
       <concept_significance>500</concept_significance>
       </concept>
 </ccs2012>
\end{CCSXML}

\ccsdesc[500]{Computing methodologies~Scene understanding}
\ccsdesc[500]{Computing methodologies~Image segmentation}

\keywords{Panoptic Narrative Grounding, Diffusion Models, Dynamic Prompting, Phrase Adapter, Multi-Level Aggregation}



\maketitle

\section{Introduction}

Given a natural language narrative caption, panoptic narrative grounding (PNG)~\cite{gonzalez2021panoptic} aims to segment things and stuff objects based on the description of noun phrases.
As an emerging task, PNG extends phrase grounding~\cite{yu2020cross} by providing more precise segmentation masks instead of bounding boxes and also shifts the grounding focus of referring expression segmentation~\cite{hu2016segmentation} from a single sentence to multiple phrases.
These characteristics of PNG enable finer vision-language understanding and open up a broad spectrum of potential applications like embodied perception~\cite{wang2023active}.

\begin{figure}[!t]
\centering
\includegraphics[width=0.85\linewidth]{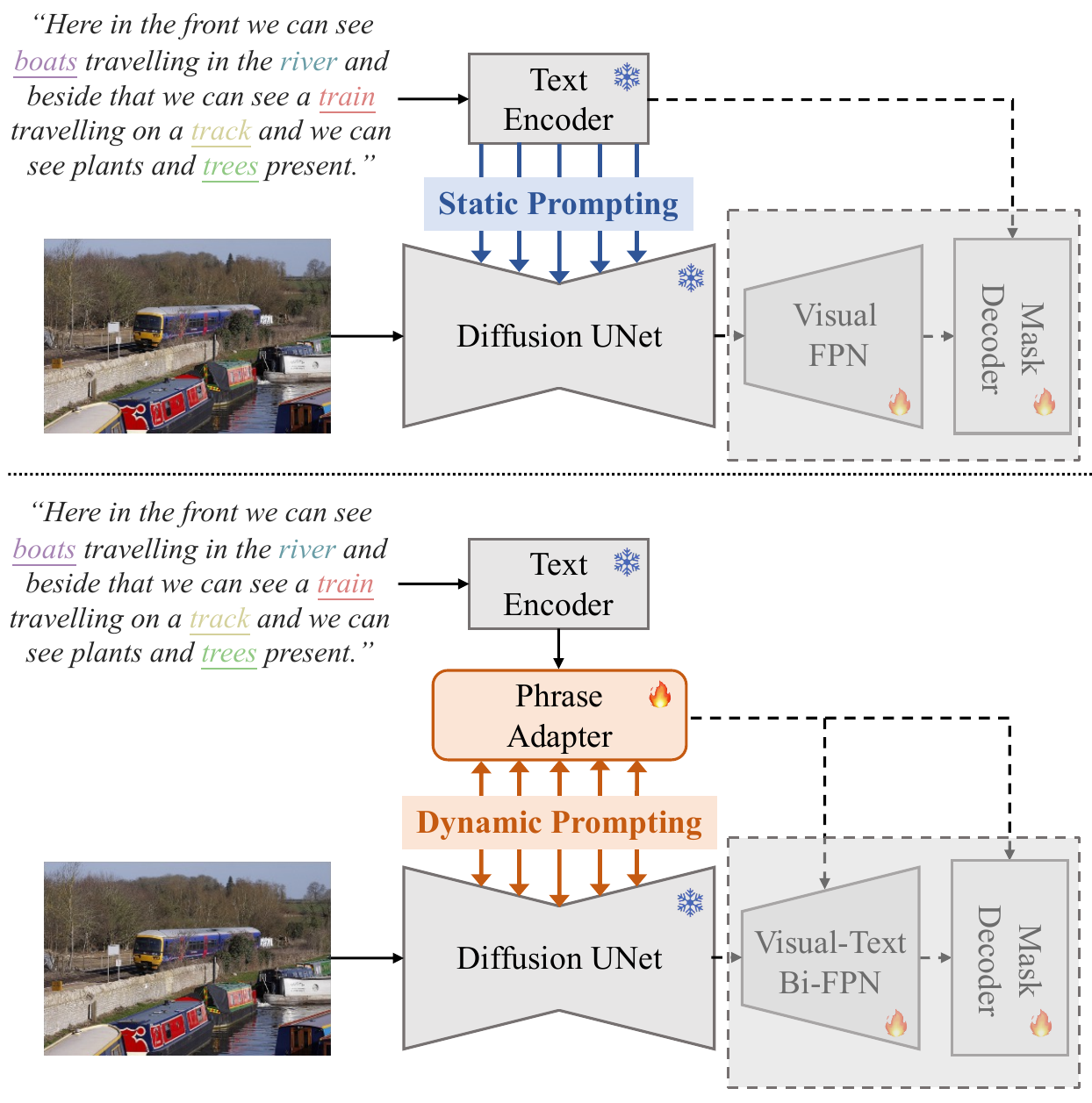}
\caption{Static prompting of frozen Diffusion models suffers from a large task gap and insufficient vision-language interaction, leading to sub-optimal generalization on the PNG task. We propose a dynamic prompting scheme via Phrase Adapters which bidirectionally update image and text features to better leverage the fine-grained image-text alignment capability of Diffusion models.}
\label{fig:intro}
\end{figure}

According to its task setting, the core target of PNG is to achieve fine-grained image-text alignment between pixels and noun phrases.
Previous methods based on the discriminative models primarily utilize two approaches to achieve this alignment.
One is to incorporate fully supervised pretraining on visual panoptic segmentation~\cite{gonzalez2021panoptic,wang2023towards,ding2022ppmn,hui2023enriching,gonzalez2023piglet} where only weak alignment between pixels and class names can be learned.
The other is to exploit the coarse-grained alignment knowledge of contrastive multimodal models like CLIP~\cite{radford2021learning,xu2023bridging} whose learning objective is global matching between image and text, leading to inaccurate pixel-level localization.

Recently, text-to-image Diffusion models~\cite{rombach2022high} have demonstrated its outstanding capability of fine-grained image-text alignment in various tasks, where local alignment between pixels and phrases in the caption can be achieved through cross-attention maps.
Moreover, the pretraining of Diffusion models does not depend on pixel-level segmentation labels.
Therefore, some pioneer works~\cite{xu2023open, wu2023diffumask, wang2023diffusion} explore frozen text-to-image Diffusion models with static text embeddings as a promising visual backbone to improve general segmentation performance.
A natural question hence arises: \textit{Could we follow these Diffusion-based segmentation methods to improve the multimodal PNG task?}.
After an in-depth analysis, we argue that the naive application of frozen Diffusion models using phrase features as static text prompts still faces serious limitations:
1) There exists a considerable gap between the pretraining and downstream tasks of the Diffusion and PNG models, making it challenging to transfer the generative knowledge from the Diffusion models to the discriminative PNG task without introducing learnable parameters in the Diffusion backbone.
2) Diffusion models contain only a unidirectional flow of information from the language domain to the visual domain, leading to images that merely capture the vague concepts of text prompts specific to the PNG task, which limits the efficacy of knowledge transfer.

To alleviate these limitations, we propose to adapt the frozen text-to-image Diffusion models to the PNG task via dynamically updated text prompts as shown in Figure~\ref{fig:intro}.
Concretely, we devise an Extractive-Injective Phrase Adapter (EIPA) which incorporates an additional adapter bypass within the Diffusion UNet to fill in the information flow from the vision domain to the language domain.
This enables bidirectional information interaction within the vision backbone, effectively transferring generative pretraining knowledge to the discriminative PNG task.
Our EIPA coordinates with the diffusion UNet's cross-attention block in terms of the same insertion position and symmetrical structure (with phrase features as the query, and image features as the key and value).
Through their collaboration, phrase features are first updated with the global context information extracted from image features and then injected back into the backbone to further update the image features with task-specific multimodal cues, ensuring sufficient generative pretrained knowledge transfer.
Additionally, we exploit the corresponding cross-attention map of each phrase from the UNet as the attention mask input to the cross-attention layers in EIPA, allowing phrases to interact with more relevant image regions.

In addition, EIPA introduces multi-level phrase features that can be aggregated with multi-level image features to combine low-level details and high-level concepts, leading to further segmentation improvement.
Therefore, we also propose to fuse these features endowed with multi-level semantics for inter-level multimodal context modeling.
Concretely, we design a Multi-Level Mutual Aggregation (MLMA) module that leverages bi-attention mechanisms~\cite{li2022grounded} to reciprocally fuse information from different levels between image and phrase features, aiming to capture image-text semantic alignments more thoroughly and enhance the quality of mask predictions.
The fused multimodal features are separately fed into a deformable attention layer~\cite{zhu2020deformable} and a self-attention layer for further refinement.
We apply a Transformer decoder~\cite{cheng2022masked} on these output features for the final mask prediction of each phrase.

The contributions of our paper are summarized as follows:
1) We propose an Extractive-Injective Phrase Adapter (EIPA) bypass within the UNet backbone to dynamically update phrase prompts with image features and inject the multimodal cues back, leading to more sufficient leverage of fine-grained image-text alignment capability of frozen text-to-image Diffusion models.
2) We also propose a Multi-Level Mutual Aggregation (MLMA) module to reciprocally fuse multi-level image and phrase features, which further refines the segmentation predictions with richer multimodal semantic information.
3) Extensive experiments on the PNG benchmark show that our method achieves new state-of-the-art performance.

\section{Related Work}

\subsection{Panoptic Narrative Grounding}
The task of panoptic narrative grounding (PNG) is first proposed by~\cite{gonzalez2021panoptic} along with a benchmark and a two-stage baseline that conducts matching between phrases and offline-produced mask proposals.
They further design an updated baseline model PiGLET~\cite{gonzalez2023piglet} where the mask embeddings of MaskFormer~\cite{cheng2021per} are used as proposals.
PPMN~\cite{ding2022ppmn} and EPNG~\cite{wang2023towards} propose one-stage end-to-end models that directly find the matched pixels for each noun phrase without relying on offline proposals, obtaining both performance improvement and speed acceleration.
DRMN~\cite{lin2023context} utilizes deformable attention to iteratively sample multi-scale pixel contexts for feature updating and alleviates the phrase-to-pixel mismatch issue.
PPO-TD~\cite{hui2023enriching} further introduces object-level context modeling and contrastive learning into the one-stage model to enhance the discriminative ability of phrase features using coupled object and pixel contexts, thereby yielding significant performance elevation.
In this paper, we propose a novel pipeline that differs greatly from previous methods, where a frozen text-to-image diffusion model is adapted by dynamically updated phrase prompts to sufficiently leverage its powerful fine-grained image-text alignment capability.

\subsection{Referring Expression Segmentation}
The goal of referring expression segmentation (RES)~\cite{hu2016segmentation} is to segment the certain object specified by the subject of a single sentence.
Early FCN-based~\cite{long2015fully} methods perform multimodal feature fusion based on diverse attention mechanisms~\cite{huang2020referring,hui2020linguistic,feng2021encoder,hui2021collaborative,liu2021cross,ding2022language}.
Some Transformer-based~\cite{vaswani2017attention} methods mainly explore the dynamic updating of language queries~\cite{ding2021vision,yang2023semantics}, multimodal fusion positions~\cite{yang2022lavt,hui2023language}, and adaptive foreground classification~\cite{kim2022restr} within the backbone.
PolyFormer~\cite{liu2023polyformer} reformulates the representation of segmenting referred objects as sequential polygon generation.
CRIS~\cite{wang2022cris} fully finetunes the discriminative CLIP model to leverage its multimodal pretrained knowledge.
BarLeRIa~\cite{wang2024barleria} proposes a bi-directional intertwined vision-language efficient tuning framework for RES.
In contrast, our method adapts the frozen text-to-image diffusion model to transfer its generative pretrained knowledge to the PNG task which grounds multiple phrases rather than a single sentence.

\begin{figure*}[!t]
\centering
\includegraphics[width=0.95\linewidth]{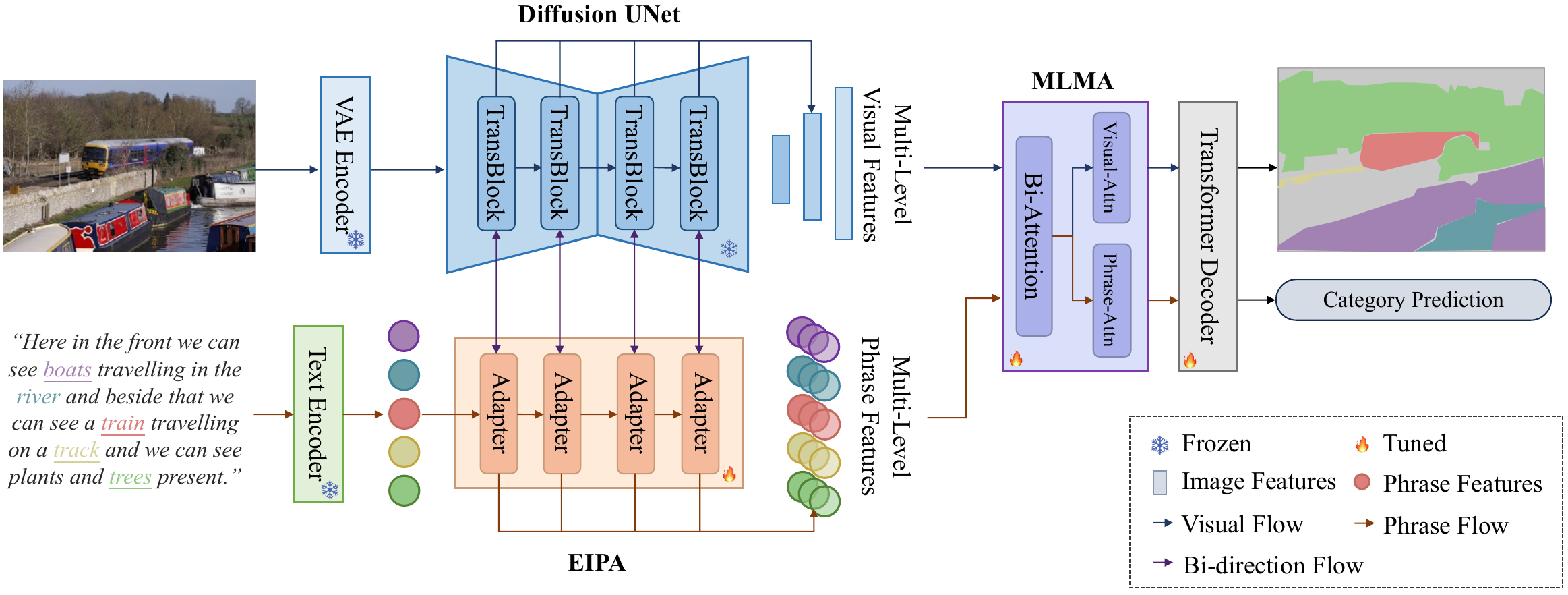}
\caption{The overall architecture of our pipeline. Input image and caption are first processed by Diffusion UNet and text encoder. An additional bypass composed of our proposed Extractive-Injective Phrase Adapter (EIPA) is introduced to update phrase features with image features, forming a bidirectional vision-language interaction. Multi-level image and phrase features obtained are further fed into our designed Multi-Level Mutual Aggregation (MLMA) module to integrate multi-level semantic information. Finally, the segmentation mask of each phrase is predicted by a Transformer decoder.}
\label{fig:method:framework}
\end{figure*}

\subsection{Diffusion Models for Segmentation}

Recently, Diffusion models have experienced notable advancements, establishing themselves as prominent generative models in contemporary research.
Observing text-to-image Diffusion models’  outstanding capability of fine-grained image-text alignment, researchers have explored their application in segmentation tasks. DiffuMask~\cite{wu2023diffumask} harnesses diffusion models to produce images and pixel-level annotations, thus training a highly effective semantic segmentation model. OVDiff~\cite{karazija2023diffusion} employs diffusion models to generate prototypes for multiple classes and subsequently matches pixel features with these prototypes during segmentation. 
DiffSegmenter~\cite{wang2023diffusion} utilizes the cross-attention map generated by diffusion models to produce masks without additional training.
VPD~\cite{zhao2023unleashing} fully finetunes the denoising UNet of diffusion models with text prompts to use its features for visual perception tasks.
ODISE~\cite{xu2023open} exploits the internal representations of frozen diffusion and CLIP models with static and implicit text embeddings to perform open-vocabulary panoptic segmentation.
In this paper, we propose an EIPA module to adapt the frozen diffusion UNet by dynamically updating the phrase prompts with image features, thereby effectively interacting between input phrase features and image features encoded by the text-to-image Diffusion models and reducing the considerable gap between Diffusion and PNG models.

\section{Method}
The overall architecture of our pipeline is illustrated in Figure~\ref{fig:method:framework}.
The input image and narrative caption are encoded by Diffusion UNet~\cite{rombach2022high} and CLIP text encoder~\cite{radford2021learning} respectively to obtain image and phrase features.
In order to sufficiently leverage the fine-grained image-text alignment capability of text-to-image Diffusion models, we propose an Extractive-Injective Phrase Adapter (EIPA) which incorporates an additional adapter bypass parallel to the Diffusion UNet to update phrase features with image features.
The updated phrase features serve as dynamic text prompts for Diffusion models to obtain better-aligned image and phrase features, thus transferring the generative pretrained knowledge to the discriminative PNG task.
The multimodal features are then fed into our devised Multi-Level Mutual Aggregation (MLMA) module to integrate multi-level semantic information from both visual and linguistic modalities.
Finally, a task decoder predicts segmentation masks for each phrase based on the refined multimodal features.

\subsection{Static Prompting for Diffusion Models}
The baseline model of our pipeline is to directly prompt the frozen text-to-image Diffusion models using static phrase features as text prompts.
Concretely, the inputs to our pipeline include an image $\mathcal{I} \in \mathbb{R}^{H^0 \times W^0 \times 3}$ and a narrative caption $\mathcal{T}$ composed of $M$ words.
For the input caption, we adopt the CLIP text encoder~\cite{radford2021learning} to extract word feature embeddings $\bm{R}_w \in \mathbb{R}^{M \times C_t}$, and conduct simple average on corresponding words to obtain phrase feature embeddings $\bm{R} \in \mathbb{R}^{N \times C_t}$, where $N$ denotes the number of phrases in the caption and $C_t$ is the channel number of phrase features.

For the input image $\mathcal{I}$, we first feed it into the VAE encoder where the image is downsampled to $1/8$ resolution with a small number of channels (\textit{i.e.}, $4$).
Then, this image feature is processed by the Diffusion UNet~\cite{ronneberger2015u} which is composed of $L$ blocks.
Typically, each UNet block contains a residual convolution block (ResBlock)~\cite{he2016deep}, a Transformer block (TransBlock)~\cite{vaswani2017attention}, and optional upsample block or downsample block.
For the clarity of presentation, we choose the core operations within the UNet, \textit{i.e.}, ResBlock and TransBlock, to represent the $l$-th UNet block with other details omitted:
\begin{equation}
\label{eq:unet_static}
    \bm{F}^{(l)} = \mathrm{UNetBlock}^{(l)}(\bm{F}^{(l-1)}, \bm{R}),
\end{equation}
where $\bm{F}^{(l-1)} \in \mathbb{R}^{H^{l-1} \times W^{l-1} \times C^{l-1}_v} and \bm{F}^{(l)} \in \mathbb{R}^{H^l \times W^l \times C^l_v}$ denote the input and output image features of the $l$-th UNet block.
The inner operations of Equation~\ref{eq:unet_static} can be formulated as follows:
\begin{equation}
\label{eq:res_static}
    \Tilde{\bm{F}}^{(l)} = \mathrm{ResBlock}^{(l)}(\bm{F}^{(l-1)}),
\end{equation}
\begin{equation}
\label{eq:trans_static}
    \bm{F}^{(l)} = \mathrm{TransBlock}^{(l)}(\Tilde{\bm{F}}^{(l)}, \bm{R}).
\end{equation}
The core operation in the Transformer block is the cross-attention layer~\cite{vaswani2017attention} where the image feature is the query and the phrase feature serves as the key and value.
Its general formulation is as follows:
\begin{equation}
    \mathrm{out} = \mathrm{CrossAttn}^{(l)}(\mathrm{query}, \mathrm{key}, \mathrm{value}),
\end{equation}
where we instantiate it with image and phrase phrases:
\begin{equation}
\label{eq:unet_ca_static}
    \begin{split}
    \Bar{\bm{F}}^{(l)}_{\text{ca}} &= \mathrm{CrossAttn}^{(l)}(\bm{F}^{(l)}_{\text{ca}}, \bm{R}, \bm{R})\\
    &= \mathrm{Softmax}\left(\frac{(\bm{F}^{(l)}_{\text{ca}}\bm{W}_{\text{q}})(\bm{R}\bm{W}_{\text{k}})^\mathrm{T}}{\sqrt{C^l_{\text{v}}}}\right)(\bm{R}\bm{W}_{\text{v}}),
    \end{split}
\end{equation}
where $\bm{F}^{(l)}_{\text{ca}}$ and $\Bar{\bm{F}}^{(l)}_{\text{ca}}$ denotes the input and output image features of the cross-attention layer.
In static prompting, we utilize the cross-attention maps between queries and keys in the UNet to obtain the final segmentation mask prediction for each phrase, which is termed the \textit{Diffusion mask head}.
For generation tasks, a noise predictor is applied to estimate the latent noise at the specific time step.
For our discriminative PNG task, we remove the noise predictor and set the time step as $1$ to avoid further information loss.

\subsection{Dynamic Prompting with Extractive-Injective Phrase Adapter}
Previous static prompting can be regarded as a direct application of Diffusion models on the PNG task in a zero-shot manner.
However, the pretraining and downstream tasks of the Diffusion and PNG models have a significant gap, and it's difficult to transfer the generative knowledge from Diffusion models to the discriminative PNG task without introducing any learnable parameters in the Diffusion backbone.
Besides, in the Diffusion models, there is only a one-way flow of information from the language domain to the image domain, resulting in the image only being able to grasp the vague linguistic concepts in the PNG task, which limits the effectiveness of knowledge transfer.
Therefore, we propose an Extractive-Injective Phrase Adapter (EIPA) which updates the phrase features with image features to fill in the information flow from the vision domain to the language domain.
Our EIPA sufficiently leverages the fine-grained image-text alignment capability of text-to-image Diffusion models by dynamically updating the text prompts.

In detail, we equip the $l$-th UNet block with a phrase adapter to construct a parallel bypass through the UNet.
Since the vision-language interaction between UNet block and phrase adapter is bidirectional, their inputs and outputs are dependent on each other, which can be formulated as:
\begin{equation}
\label{eq:unet_dynamic}
    \bm{F}^{(l)}, \bm{F}^{(l)}_\text{itm} = \mathrm{UNetBlock}^{(l)}(\bm{F}^{(l-1)}, \bm{R}^{(l)}_{\text{itm}}),
\end{equation}
\begin{equation}
\label{eq:adap_overall}
    \bm{R}^{(l)}, \bm{R}^{(l)}_\text{itm} = \mathrm{PhraseAdapter}^{(l)}(\bm{F}^{(l)}_\text{itm}, \bm{R}^{(l-1)}),
\end{equation}
where $\bm{F}^{(l)}_\text{itm}$ and $\bm{R}^{(l)}_\text{itm}$ are the intermediate outputs from the self-attention layers in the Transformer blocks of UNet block and Phrase adapter.
From the comparison between Equation~\ref{eq:unet_static} and Equation~\ref{eq:unet_dynamic}, we can observe that the phrase feature fed into the UNet block is iteratively updated.
For the $l$-th phrase adapter of Equation~\ref{eq:adap_overall}, we can expand its inner operations as follows:
\begin{equation}
\label{eq:adap_sa}
    \bm{R}^{(l)}_\text{itm} = \mathrm{SelfAttn}^{(l)}(\bm{R}^{(l-1)}) + \bm{R}^{(l-1)},
\end{equation}
\begin{equation}
\label{eq:adap_ca}
    \Bar{\bm{R}}^{(l)} = \mathrm{CrossAttn}^{(l)}(\bm{R}^{(l)}_\text{itm}, \bm{F}^{(l)}_\text{itm}, \bm{F}^{(l)}_\text{itm}) + \bm{R}^{(l)}_\text{itm},
\end{equation}
\begin{equation}
\label{eq:adap_ffn}
    \bm{R}^{(l)} = \mathrm{FFN}^{(l)}(\Bar{\bm{R}}^{(l)}) + \Bar{\bm{R}}^{(l)}.
\end{equation}

\begin{figure}[!t]
  \begin{center}
     \includegraphics[width=0.9\linewidth]{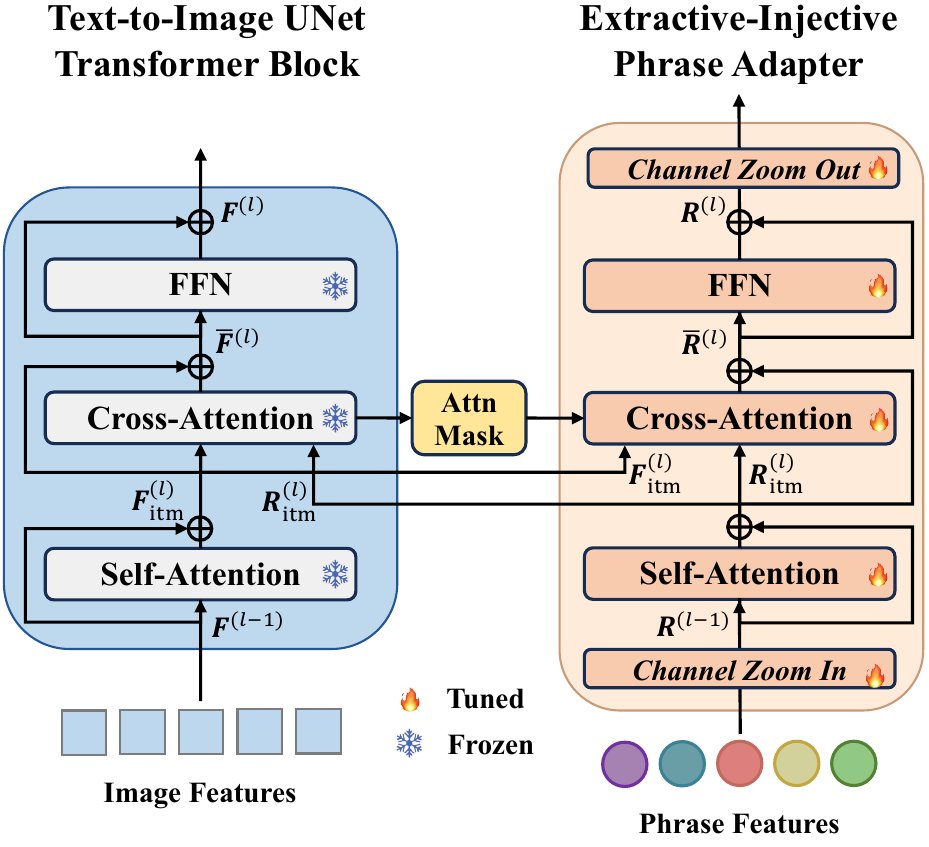}
  \end{center}
     \caption{The detailed structure of Extractive-Injective Phrase Adapter (EIPA). Feature dimensions in adapters are zoomed in and out to reduce the number of tuned parameters.}
  \label{fig:method:adapter}
\end{figure}

The channel numbers of input and output features of the phrase adapter are zoomed in and zoomed out to reduce parameters, which are omitted here. Moreover, we exploit the segmentation mask predictions from the Diffusion mask head (see discussion of Equation~\ref{eq:unet_ca_static}) to serve as the attention masks to the cross-attention layer for phrase queries in our EIPA.
Thus, the attended region of each phrase can be restricted to predicted foreground areas for noise reduction following~\cite{cheng2022masked}.
We also integrate all the cross-attention maps in our EIPA to predict the segmentation mask for each phrase, which is termed as the \textit{adapter mask head} to provide another intermediate supervision.
The inner operations of the Transformer block in the $l$-th UNet block can be expanded similarly as Equation~\ref{eq:adap_sa}-\ref{eq:adap_ffn}, where the cross-attention layer uses $\bm{F}^{(l)}_\text{itm}$ as the query and $\bm{R}^{(l)}_\text{itm}$ as the key and value.
The computation process of our proposed EIPA is also shown in Figure~\ref{fig:method:adapter} for more details.
Benefiting from the bidirectional vision-language interaction, our EIPA extracts global contexts from image features to dynamically update the text prompts and then injects the task-specific multimodal cues back into the image features for more sufficient knowledge transfer.

\subsection{Multi-Level Mutual Aggregation}
Since EIPA introduces multi-level phrase features that can be aggregated with multi-level image features to combine low-level details and high-level concepts, we also propose to aggregate them for inter-level multimodal context modeling.
Therefore, we design a Multi-Level Mutual Aggregation (MLMA) module that exploits bi-attention~\cite{li2022grounded} to reciprocally fuse multi-level information of image and phrase features in order to more comprehensively model image-text semantic correspondences for better segmentation quality.

Concretely, we obtain three levels of image features from different blocks in the Diffusion UNet and project their feature channels to the same number, which are denoted as $\{\bm{F}_\text{i}\}_\text{i=3}^5 \in \mathbb{R}^{H^i \times W^i \times C_\text{m}}$, $H^i = \frac{H^0}{2^i}, W^i = \frac{W^0}{2^i}$.
The resolutions of $\{\bm{F}_\text{i}\}_\text{i=3}^5$ are $1/8, 1/16, 1/32$ of the input image.
Accordingly, we obtain multi-level phrase features $\{\bm{R}_\text{i}\}_\text{i=3}^5 \in \mathbb{R}^{N \times C_\text{m}}$ from different blocks in our EIPA which contain semantic information relevant to the corresponding levels of image features.
Different from cross-attention, bi-attention~\cite{li2022grounded} computes the attention map once and applies Softmax normalization on the dimension of pixel number or phrase number respectively, then multiplies with image or phrase features for bilateral fusion.
Our MLMA utilizes the concatenation of $\{\bm{R}_\text{i}\}_\text{i=3}^5$ on the phrase number dimension as the query and the concatenation of $\{\bm{F}_\text{i}\}_\text{i=3}^5$ on the pixel number dimension as the key to compute the bi-attention map.
The value is the concatenation of $\{\bm{F}_\text{i}\}_\text{i=3}^5$ or $\{\bm{R}_\text{i}\}_\text{i=3}^5$ for multi-level information aggregation in two directions respectively.
After bi-attention, we feed multi-level image features into a deformable attention layer~\cite{zhu2020deformable} and multi-level phrase features into a self-attention layer for intra-modal refinement.

In addition, image feature after the UNet is input to the VAE decoder to obtain feature $\Hat{\bm{F}}_2$ of $1/4$ resolution of the original image.
The image feature $\Hat{\bm{F}}_3$ of $1/8$ resolution output by our MLMA is further upsampled and fused with $\Hat{\bm{F}}_2$ to obtain the final mask feature $\bm{F}_m$.
The last level of output phrase features $\Hat{\bm{R}}_5$ and multi-level image features $\{\Hat{\bm{F}}_\text{i}\}_\text{i=3}^5$ from our MLMA are further fed into a Transformer decoder~\cite{cheng2022masked} to yield final phrase queries.
The segmentation mask for each phrase is predicted by matrix multiplication between each phrase query and mask feature $\bm{F}_\text{m}$, which is termed as the \textit{decoder mask head} to provide final supervision.

\subsection{Loss Functions}
As mentioned before, our model contains three mask heads for loss supervision.
The Diffusion mask head fuses cross-attention maps after Softmax (Equation~\ref{eq:unet_ca_static}) in all UNet blocks with weighted summation to obtain predicted segmentation masks $\bm{Y}^{\mathrm{dif}} \in \mathbb{R}^{N \times H^0 \times W^0}$.
Given the ground-truth segmentation masks $\bm{Y}^* \in \mathbb{R}^{N \times H^0 \times W^0}$, we apply cross-entropy (CE) loss between $\bm{Y}^{\mathrm{dif}}$ and $\bm{Y}^*$:
\begin{equation}
    \mathcal{L}_{\mathrm{mask}}^{\mathrm{dif}} = \sum_{j=1}^N\mathcal{L}_{\text{ce}}(\bm{y}^{\mathrm{dif}}_\text{j}, \bm{y}^*_\text{j}) = \frac{1}{NH^0W^0}\sum_{j=1}^N\sum_{i=1}^{H^0W^0}-y^*_\text{ij}\mathrm{log}p(y^{\mathrm{dif}}_\text{ij}),
\end{equation}
where $\bm{y}^{\mathrm{dif}}_\text{j} \in \mathbb{R}^{H^0 \times W^0}$ and $\bm{y}^*_\text{j} \in \mathbb{R}^{H^0 \times W^0}$ is the segmentation prediction of the Diffusion mask head and the ground-truth for each phrase respectively.
For the adapter mask head and decoder mask head, we adopt the mask classification loss from Mask2Former~\cite{cheng2022masked}:
\begin{equation}
    \mathcal{L}_{\mathrm{mask\text{-}cls}}^{\mathrm{ada}} = \sum_{j=1}^N\left[-{\mathrm{log}}p(c^*_\text{j}) + \mathcal{L}_{\mathrm{mask}}(\bm{y}^{\mathrm{ada}}_\text{j}, \bm{y}^*_\text{j})\right],
\end{equation}
\begin{equation}
    \mathcal{L}_{\mathrm{mask\text{-}cls}}^{\mathrm{dec}} = \sum_{j=1}^N\left[-{\mathrm{log}}p(c^*_\text{j}) + \mathcal{L}_{\mathrm{mask}}(\bm{y}^{\mathrm{dec}}_\text{j}, \bm{y}^*_\text{j})\right],
\end{equation}
where $c^*_\text{j}$ is the ground-truth category of each phrase to utilize category priors in phrases, and $\mathcal{L}_{\rm{mask}}$ is the combination of binary cross entropy (BCE) loss and Dice loss~\cite{sudre2017generalised}.
We apply $\mathcal{L}_{\rm{mask\text{-}cls}}^{\mathrm{ada}}$ in each phrase adapter block of our EIPA.
The total loss of our model is then computed as the sum of the above three individual losses:
\begin{equation}
    \mathcal{L} = \mathcal{L}_{\mathrm{mask}}^{\mathrm{dif}} + \mathcal{L}_{\mathrm{mask\text{-}cls}}^{\mathrm{ada}} + \mathcal{L}_{\mathrm{mask\text{-}cls}}^{\mathrm{dec}}.
\end{equation}

\begin{table*}[t]
  \centering
    \label{tab:things_stuff_results}
    {\resizebox{0.9\linewidth}{!}{
    \begin{tabular}{r|c|c|c||c|c|c|c|c}
    \hline\thickhline
    & & & & \multicolumn{5}{c}{Average Recall} \\
    \multirow{-2}*{Method}  & \multirow{-2}*{Text Encoder}  & \multirow{-2}*{Diffusion}  & \multirow{-2}*{P.S. Pretrain} & \multicolumn{1}{c}{overall} & \multicolumn{1}{c}{things} & \multicolumn{1}{c}{stuff} & \multicolumn{1}{c}{singulars} & \multicolumn{1}{c}{plurals} \\ \hline\hline
    EPNG~\shortcite{wang2023towards}\tiny{AAAI23}       & BERT~\shortcite{devlin2018bert} &\ding{55} &\ding{55}  & 49.7 & 45.6 & 55.5 & 50.2 & 45.1 \\
    MCN~\shortcite{luo2020multi}\tiny{CVPR20}           & BERT~\shortcite{devlin2018bert} &\ding{55} &\ding{51}  & 54.2 & 48.6 & 61.4 & 56.6 & 38.8 \\
    PNG~\shortcite{gonzalez2021panoptic}\tiny{ICCV21}   & BERT~\shortcite{devlin2018bert} &\ding{55} &\ding{51}  & 55.4 & 56.2 & 54.3 & 56.2 & 48.8 \\
    PPMN~\shortcite{ding2022ppmn}\tiny{ACMMM22}         & BERT~\shortcite{devlin2018bert} &\ding{55} &\ding{55}  & 56.7 & 53.4 & 61.1 & 57.4 & 49.8 \\
    EPNG~\shortcite{wang2023towards}\tiny{AAAI23}       & BERT~\shortcite{devlin2018bert} &\ding{55} &\ding{51}  & 58.0 & 54.8 & 62.4 & 58.6 & 52.1 \\
    PPMN~\shortcite{ding2022ppmn}\tiny{ACMMM22}         & BERT~\shortcite{devlin2018bert} &\ding{55} &\ding{51}   & 59.4 & 57.2 & 62.5 & 60.0 & 54.0 \\
    ODISE~\shortcite{xu2023open}\tiny{CVPR23}           & CLIP~\shortcite{radford2021learning} &\ding{51} &\ding{55}   & 61.0 & 57.0 & 66.6 & 61.7 & 54.8 \\
    NICE~\shortcite{wang2023nice}\tiny{arXiv}           & BERT~\shortcite{devlin2018bert} &\ding{55} &\ding{51}   & 62.3 & 60.2 & 65.3 & 63.1 & 55.2 \\
    DRMN~\shortcite{lin2023context}\tiny{ICDM23}        & BERT~\shortcite{devlin2018bert} &\ding{55} &\ding{51}  & 62.9 & 60.3 & 66.4 & 63.6 & 56.7 \\
    ODISE~\shortcite{xu2023open}\tiny{CVPR23}           & CLIP~\shortcite{radford2021learning} &\ding{51} &\ding{51}   & 63.1 & 59.6 & 68.0 & 64.0 & 55.1 \\
    PiGLET~\shortcite{gonzalez2023piglet}\tiny{TPAMI23} & BERT~\shortcite{devlin2018bert}/~RoBERTa~\shortcite{liu2019roberta}/~CLIP~\shortcite{radford2021learning}/~GPT2~\shortcite{radford2019language} &\ding{55} &\ding{51}   & 65.9 & 64.0 & 68.6 & 67.2 & 54.5 \\
    PPO-TD~\shortcite{hui2023enriching}\tiny{IJCAI23}   & BERT~\shortcite{devlin2018bert}/~CLIP~\shortcite{radford2021learning}/~T5~\shortcite{raffel2020exploring} &\ding{55} &\ding{51}  & 66.1 & 63.3 & 69.8 & 66.9 & 58.6 \\
    \hline\hline
    Ours & CLIP~\shortcite{radford2021learning}  & \ding{51} & \ding{55} & 64.5 & 60.8 & 69.7 & 65.5 & 55.6 \\ \hline
    Ours  & CLIP~\shortcite{radford2021learning} & \ding{51}  & \ding{51} & \textbf{67.1} & \textbf{64.3} & \textbf{71.0}  & \textbf{67.9} & \textbf{60.0} \\
    \hline
    \end{tabular}
    }}
\caption{Comparison with previous state-of-the-art methods on the PNG benchmark, disaggregated into things and stuff categories, and singulars and plurals noun phrases. ``P.S. Pretrain'' denotes visual panoptic segmentation pretraining on COCO. The highest performances are reported among different text encoders.}
\label{tab:sota_results}
\end{table*}

\begin{figure*}[!htbp]
   \centering
   \medskip
   \begin{subfigure}[t]{.3\linewidth}
  \centering
  \resizebox{0.95\linewidth}{!}{%
  \begin{tikzpicture}[/pgfplots/width=1.45\linewidth, /pgfplots/height=1.45\linewidth]
    \begin{axis}[% Axis labels
                 ymin=0,ymax=1,xmin=0,xmax=1,
    			 % Axis labels
        		 xlabel=IoU,
        		 ylabel=Recall@IoU,
         		 xlabel shift={-2pt},
        		 ylabel shift={-3pt},
         		 % General appearance
		         font=\small,
		         axis equal image=true,
		         enlargelimits=false,
		         clip=true,
		         % Grids
        	     grid style=solid, grid=both,
                 major grid style={white!85!black},
        		 minor grid style={white!95!black},
		 		 xtick={0,0.1,...,1.1},
                 xticklabels={0,.1,.2,.3,.4,.5,.6,.7,.8,.9,1},
        		 ytick={0,0.1,...,1.1},
                 yticklabels={0,.1,.2,.3,.4,.5,.6,.7,.8,.9,1},
         		 minor xtick={0,0.02,...,1},
		         minor ytick={0,0.02,...,1},
        		 % Legend
        		 legend style={at={(0.05,0.05)},
                 		       anchor=south west},
                 legend cell align={left}]
    \addplot+[red,solid,mark=none,ultra thick] table[x=IoU,y=overall]{figs/sota.txt};
    \addlegendentry{EIPA+MLMA}
    \addplot+[green,solid,mark=none,ultra thick] table[x=IoU,y=overall]{figs/mlma.txt};
    \addlegendentry{MLMA only}
    \addplot+[blue,solid,mark=none,ultra thick] table[x=IoU,y=overall]{figs/eipa.txt};
    \addlegendentry{EIPA only}
    \addplot+[olive,solid,mark=none,ultra thick] table[x=IoU,y=overall]{figs/baseline.txt};
    \addlegendentry{Static Baseline}
    \end{axis}
\end{tikzpicture}}
  \subcaption{Overall performance}
  \label{fig:overall}
\end{subfigure}
\begin{subfigure}[t]{.3\linewidth}
  \centering
    \resizebox{0.95\linewidth}{!}{%
  \begin{tikzpicture}[/pgfplots/width=1.45\linewidth, /pgfplots/height=1.45\linewidth]
    \begin{axis}[% Axis labels
                 ymin=0,ymax=1,xmin=0,xmax=1,
    			 % Axis labels
        		 xlabel=IoU,
        		 ylabel=Recall@IoU,
         		 xlabel shift={-2pt},
        		 ylabel shift={-3pt},
         		 % General appearance
		         font=\small,
		         axis equal image=true,
		         enlargelimits=false,
		         clip=true,
		         % Grids
        	     grid style=solid, grid=both,
                 major grid style={white!85!black},
        		 minor grid style={white!95!black},
		 		 xtick={0,0.1,...,1.1},
                 xticklabels={0,.1,.2,.3,.4,.5,.6,.7,.8,.9,1},
        		 ytick={0,0.1,...,1.1},
                 yticklabels={0,.1,.2,.3,.4,.5,.6,.7,.8,.9,1},
         		 minor xtick={0,0.02,...,1},
		         minor ytick={0,0.02,...,1},
        		 % Legend
        		 legend style={at={(0.05,0.05)},
                 		       anchor=south west},
                 legend cell align={left}]
    \addplot+[red,dashed,mark=none,ultra thick] table[x=IoU,y=things]{figs/sota.txt};
    \addlegendentry{EIPA+MLMA(Things)}
    \addplot+[green,dashed,mark=none,ultra thick] table[x=IoU,y=things]{figs/mlma.txt};
    \addlegendentry{MLMA only(Things)}
    \addplot+[blue,dashed,mark=none,ultra thick] table[x=IoU,y=things]{figs/eipa.txt};
    \addlegendentry{EIPA only(Things)}
    \addplot+[olive,dashed,mark=none,ultra thick] table[x=IoU,y=things]{figs/baseline.txt};
    \addlegendentry{Static Baseline(Things)}
    \addplot+[red,solid,mark=none,ultra thick] table[x=IoU,y=stuff]{figs/sota.txt};
    \addlegendentry{EIPA+MLMA(Stuff)}
    \addplot+[green,solid,mark=none,ultra thick] table[x=IoU,y=stuff]{figs/mlma.txt};
    \addlegendentry{MLMA only(Stuff)}
    \addplot+[blue,solid,mark=none,ultra thick] table[x=IoU,y=stuff]{figs/eipa.txt};
    \addlegendentry{EIPA only(Stuff)}
    \addplot+[olive,solid,mark=none,ultra thick] table[x=IoU,y=stuff]{figs/baseline.txt};
    \addlegendentry{Static Baseline(Stuff)}
    \end{axis}
\end{tikzpicture}}
  \subcaption{Things and stuff categories}
  \label{fig:things_stuff}
\end{subfigure}
\begin{subfigure}[t]{.3\linewidth}
  \centering
      \resizebox{0.95\linewidth}{!}{%
  \begin{tikzpicture}[/pgfplots/width=1.45\linewidth, /pgfplots/height=1.45\linewidth]
    \begin{axis}[% Axis labels
                 ymin=0,ymax=1,xmin=0,xmax=1,
    			 % Axis labels
        		 xlabel=IoU,
        		 ylabel=Recall@IoU,
         		 xlabel shift={-2pt},
        		 ylabel shift={-3pt},
         		 % General appearance
		         font=\small,
		         axis equal image=true,
		         enlargelimits=false,
		         clip=true,
		         % Grids
        	     grid style=solid, grid=both,
                 major grid style={white!85!black},
        		 minor grid style={white!95!black},
		 		 xtick={0,0.1,...,1.1},
                 xticklabels={0,.1,.2,.3,.4,.5,.6,.7,.8,.9,1},
        		 ytick={0,0.1,...,1.1},
                 yticklabels={0,.1,.2,.3,.4,.5,.6,.7,.8,.9,1},
         		 minor xtick={0,0.02,...,1},
		         minor ytick={0,0.02,...,1},
        		 % Legend
        		 legend style={at={(0.05,0.05)},
                 		       anchor=south west},
                 legend cell align={left}]
    \addplot+[red,dashed,mark=none,ultra thick] table[x=IoU,y=singulars]{figs/sota.txt};
    \addlegendentry{EIPA+MLMA(Singulars)}
    \addplot+[green,dashed,mark=none,ultra thick] table[x=IoU,y=singulars]{figs/mlma.txt};
    \addlegendentry{MLMA only(Singulars)}
    \addplot+[blue,dashed,mark=none,ultra thick] table[x=IoU,y=singulars]{figs/eipa.txt};
    \addlegendentry{EIPA only(Singulars)}
    \addplot+[olive,dashed,mark=none,ultra thick] table[x=IoU,y=singulars]{figs/baseline.txt};
    \addlegendentry{Static Baseline(Singulars)}
    \addplot+[red,solid,mark=none,ultra thick] table[x=IoU,y=plurals]{figs/sota.txt};
    \addlegendentry{EIPA+MLMA(Plurals)}
    \addplot+[green,solid,mark=none,ultra thick] table[x=IoU,y=plurals]{figs/mlma.txt};
    \addlegendentry{MLMA only(Plurals)}
    \addplot+[blue,solid,mark=none,ultra thick] table[x=IoU,y=plurals]{figs/eipa.txt};
    \addlegendentry{EIPA only(Plurals)}
    \addplot+[olive,solid,mark=none,ultra thick] table[x=IoU,y=plurals]{figs/baseline.txt};
    \addlegendentry{Static Baseline(Plurals)}
    \end{axis}
\end{tikzpicture}}
  \subcaption{Singulars and plurals}
  \label{fig:singulars_plurals}
\end{subfigure}
   \caption{Average Recall curves of our model ablations in Table~\ref{tab:ablation:components}, (a) comparing four component analysis ablations, disaggregated into (b) things and stuff categories, and (c) singulars and plurals noun phrases.}
  \label{figs:average_recall_curve}
\end{figure*}
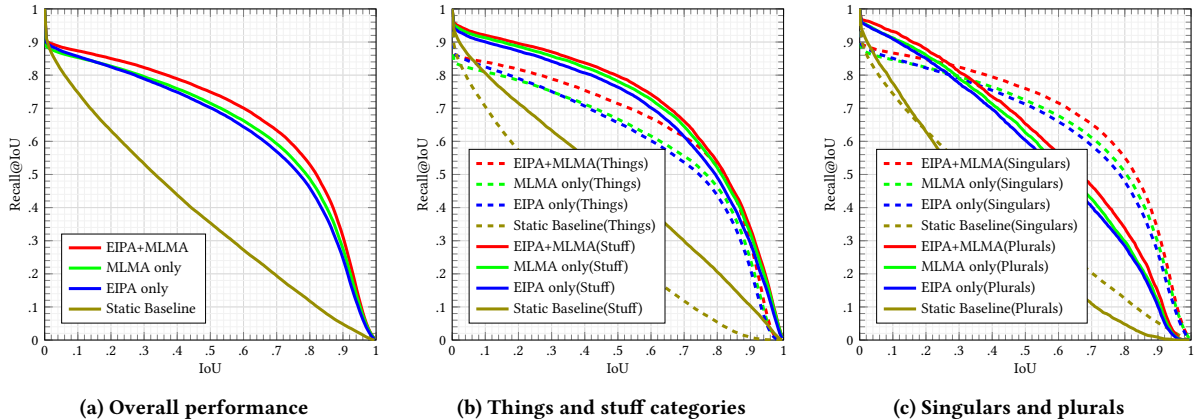

\section{Experiments}
\subsection{Dataset and Evaluation Metrics}
Following prior works~\cite{ding2022ppmn,hui2023enriching}, the dataset we used to perform experiments is the Panoptic Narrative Grounding (PNG) benchmark~\cite{gonzalez2021panoptic}, which is built by combining the narrative captions annotated in the Localized Narratives dataset~\cite{pont2020connecting} with the panoptic segmentation annotated in the COCO dataset~\cite{lin2014microsoft}.
In the PNG benchmark, there is a total number of 726,445 noun phrases that are matched with 741,697 segmentation masks to form caption-image annotation pairs.
For each caption, an average of 11.3 noun phrases are included with 5.1 of them requiring grounding.

In terms of metrics, Average Recall (AR) is adopted for model evaluation.
For each phrase, if the Intersection over Union (IoU) between the predicted segmentation and the ground-truth segmentation is above a certain threshold, then this prediction will be regarded as a true positive.
By enumerating recall values at different IoU thresholds, we can draw a recall curve where the area under the curve is determined to be the Average Recall.
For each plural phrase, its ground-truth segmentation masks are combined as a single mask, and so are the predicted masks.
Then, the IoU is computed between these two combined masks.

\subsection{Implementation Details}
Our method is implemented using PyTorch, with the input image resized to $1024 \times 1024$.
We adopt the Stable Diffusion model pretrained on a subset of the LAION dataset as our text-to-image Diffusion model.
The time step used for the diffusion process is set to $t = 0$. We employ CLIP for text encoding.
In EIPA, the phrase is default zoomed in to the dimension of 64 and our adapters have a total of 3.37M parameters.
Our design of the mask decoder follows Mask2Former ~\cite{cheng2022masked} architecture.
The maximum length of input captions is restricted to 230 words, with a requirement for grounding up to $N = 30$ noun phrases.
We utilize AdamW as the optimizer with a learning rate of $1e^{-4}$ and train our model with a batch size of 16 for 180K iterations on 4 NVIDIA A100 GPUs.
The parameters of the text encoder, Diffusion UNet, and VAE encoder remain fixed during training.
To further enhance the quality of generated masks, we leverage panoptic pretraining on MSCOCO, aligning with previous discriminative methods.
Our Transformer decoder and visual attention in MLMA utilize pretrained parameters from ODISE~\cite{xu2023open}.

\subsection{Comparison with State-of-the-Art Methods}
We compare our proposed method with previous state-of-the-art methods on the PNG benchmark.
Table~\ref{tab:sota_results} summarizes the comparison results on the overall set and things/stuff/singulars/plurals subsets of the PNG benchmark.
Compared to previous methods that relied on discriminative pretraining for panoptic segmentation, 
our approach with additional generative pretraining achieves state-of-the-art performance, demonstrating the utility of fine-grained image-text alignment capabilities inherent in large-scale text-to-image diffusion models.
By introducing phrase adapters, the generative pretraining knowledge is successfully transferred to the discriminative PNG task, resulting in superior performance.
Compared to PPO-TD~\cite{hui2023enriching} and PiGLET~\cite{gonzalez2023piglet} which use pretrained Mask2Former~\cite{cheng2022masked} as visual backbones, our results indicate that using a pretrained diffusion model as the visual backbone can provide the decoder and segmentation head with visual features capturing the detailed correspondences between objects and text, proving the significant application potential of generative models in the PNG task.
ODISE~\cite{xu2023open} also utilizes the frozen Diffusion model as the visual backbone for open-vocabulary segmentation where only uni-directional interaction from implicit text embedding to image features exists, and it adopts a mask decoder pretrained on panoptic segmentation data as well.
We reimplement ODISE for the PNG task and our method without panoptic segmentation pretraining achieves higher performance than ODISE, which demonstrates the efficacy of adapting large text-to-image Diffusion models with dynamic prompting and multi-level mutual aggregation.

\subsection{Ablation Studies}
We also conduct ablation studies on the PNG benchmark to verify the effectiveness of our network designs.

\begin{table}[!htbp]
    \centering
    {
    \resizebox{1.0\linewidth}{!}{
    \begin{tabular}{c|c||c|c|c|c|c}
    \hline\thickhline
     &  & \multicolumn{5}{c}{Average Recall} \\
    \multirow{-2}*{EIPA} &\multirow{-2}*{MLMA}
      &\multicolumn{1}{c}{overall} & \multicolumn{1}{c}{things} & \multicolumn{1}{c}{stuff} & \multicolumn{1}{c}{singulars} & \multicolumn{1}{c}{plurals} \\ \hline\hline 
                &            & 37.8 & 31.6 & 46.3 & 38.2 & 33.4 \\
     \checkmark &            & 62.9 & 59.4 & 67.7 & 63.6 & 56.0 \\
                & \checkmark & 64.1 & 60.0 & 70.0 & 64.9 & 57.4 \\
     \checkmark & \checkmark & \textbf{67.1} & \textbf{64.3} & \textbf{71.0} & \textbf{67.9} &\textbf{60.0} \\
     \hline
    \end{tabular}}
    }
    \caption{Verifying the effectiveness of components in our method. MLMA is added with the Transformer decoder and contains panoptic segmentation pretraining.}
    \label{tab:ablation:components}
\end{table}

\begin{figure*}[t] 
\centering
\includegraphics[width=0.8\linewidth]{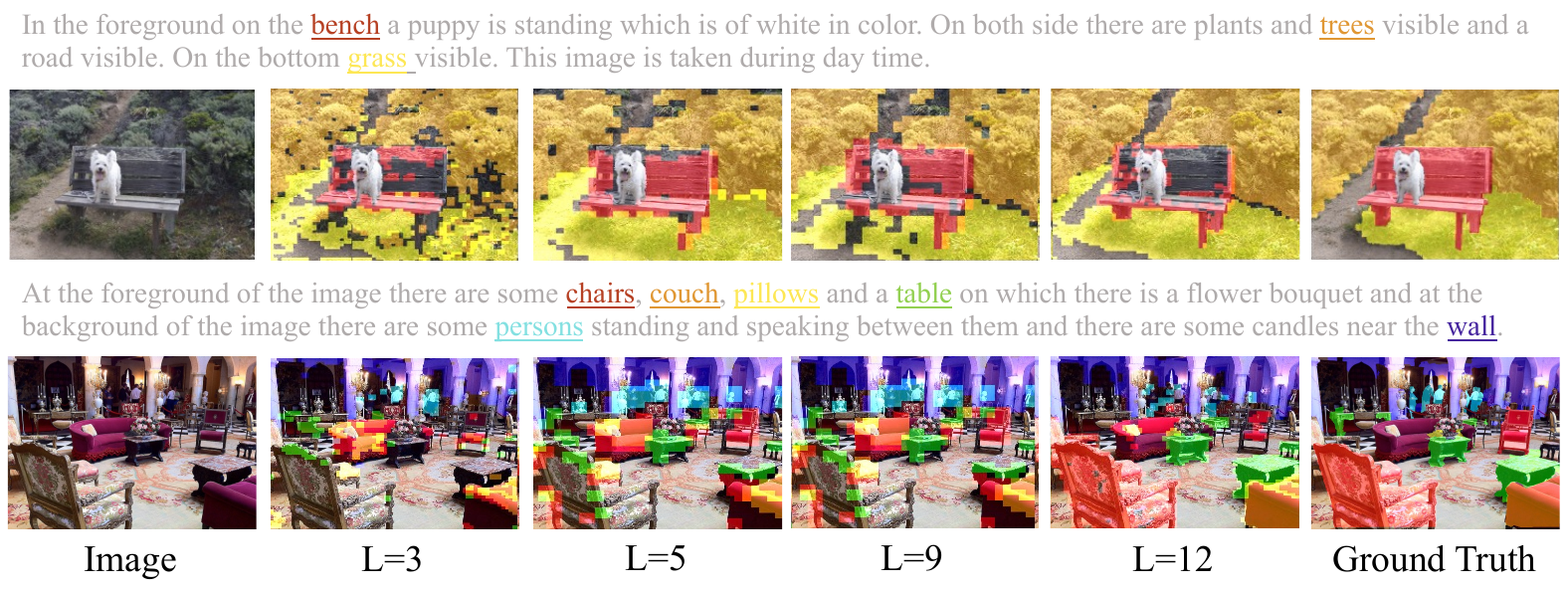}
\caption{Visualization of cross-attention maps in different layers ($L$) of our EIPA. We assign the most matched phrase label to each pixel to illustrate the overall effect of cross-attention.}
\label{fig:vis:adapter_attn}
\end{figure*}

\begin{figure*}[!htbp] 
\centering
\includegraphics[width=0.8\linewidth]{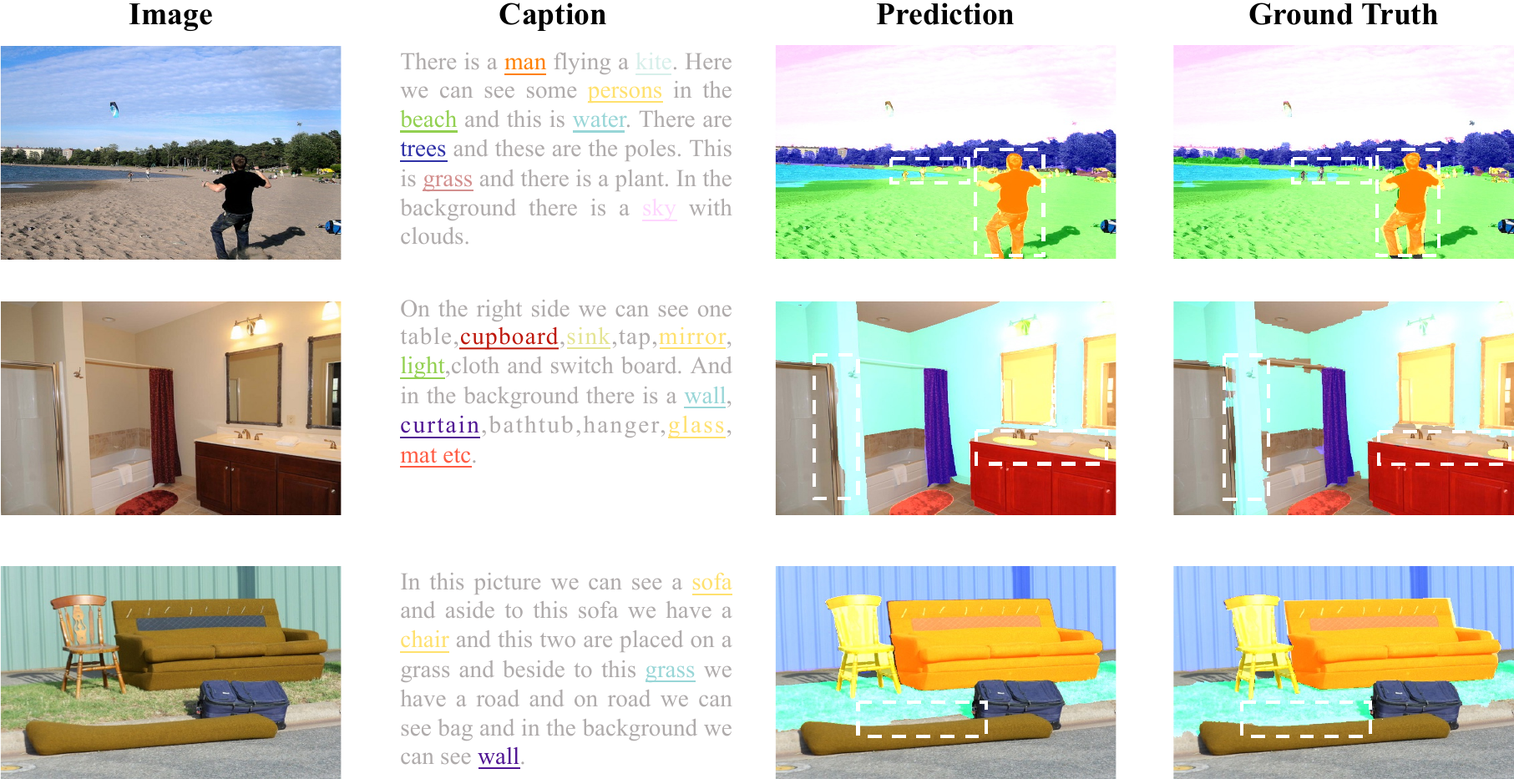}
\caption{Qualitative comparison between our method's predictions and ground-truth annotations.}
\label{fig:vis:pred_vs_gt}
\end{figure*}

\textbf{Component Analysis.}
In Table~\ref{tab:ablation:components}, we analyze the effects of our proposed modules.
The 1-st row denotes our baseline model where the segmentation masks are generated from both cross-attention maps in the Diffusion UNet and the inner product between image and phrase features.
Compared to the 1-st row, the introduction of our proposed EIPA into the Diffusion model in the 2-nd row leads to a noticeable improvement in segmentation performance.
This demonstrates that incorporating learnable parameters for bidirectional interaction between image and phrase features can transfer the generative pre-training knowledge from the Diffusion model more effectively, thus making fuller use of the fine-grained image-text alignment capability of large-scale text-to-image Diffusion models.
Incorporating our proposed MLMA module separately or together with EIPA can yield performance elevation as shown in the last two rows, which suggests that aggregating multimodal semantic information at multiple levels can lead to a more comprehensive understanding of visual scenes.

In addition, we present the recall curves of different ablation models from Table~\ref{tab:ablation:components} in Figure~\ref{figs:average_recall_curve} where performances on all subsets are also shown.
On most IoU thresholds, the curves of our models are higher than that of the baseline model, which indicates our proposed components are beneficial for identifying more referred targets as well as achieving precise segmentation results.

\begin{table}[!htbp] 
    \centering
    {
    \resizebox{1.0\linewidth}{!}{
    \begin{tabular}{c||c|c|c|c|c}
    \hline\thickhline
    & \multicolumn{5}{c}{Average Recall} \\
    \multirow{-2}*{Adapter~Position} & \multicolumn{1}{c}{overall} & \multicolumn{1}{c}{things} & \multicolumn{1}{c}{stuff} & \multicolumn{1}{c}{sigulars} & \multicolumn{1}{c}{plurals} \\ \hline\hline
    No-Adapter & 63.1 & 59.6 & 68.0 & 64.0 & 55.1 \\
    Encoder & 65.3 & 62.1 & 69.8 & 66.2 & 56.8 \\
    Decoder & 63.2 & 59.4 & 68.4 & 64.0 & 55.8 \\
    Encoder-Decoder & \textbf{66.0} & \textbf{62.7} & \textbf{70.5} & \textbf{66.9} & \textbf{57.6} \\\hline
    \end{tabular}}
    }
    \caption{Results of inserting positions of EIPA in the UNet.}
    \label{tab:ablation:eipa_position}
\end{table}

\textbf{Positions of EIPA layers.}
In Table~\ref{tab:ablation:eipa_position}, we analyze the effects of inserting EIPA at different positions of the Diffusion UNet. Visual Deformable Attention and Transformer Decoder is used without incorporating Bi-Attention and Text Self-Attention in the MLMA module.
Given that UNet is divided into encoder and decoder parts, EIPA was separately inserted into the encoder and decoder of UNet for ablation studies.
The experimental results reveal that inserting the phrase adapters into either the encoder or decoder of UNet yields better performance than not inserting it at all, with a notable improvement observed when inserted into the encoder.
Moreover, inserting the phrase adapters into both the encoder and decoder parts of UNet leads to further performance enhancements, which indicates that the more layers of UNet are adapted with visual features, the better the segmentation performance typically is on the downstream PNG task.

\begin{table}[!htbp]
    \centering
    {
    \resizebox{1.0\linewidth}{!}{
    \begin{tabular}{c|c|c||c|c|c|c|c}
    \hline\thickhline
     &  & & \multicolumn{5}{c}{Average Recall} \\
    \multirow{-2}*{Deform-Attn} & \multirow{-2}*{Bi-Attn} & \multirow{-2}*{Text-Attn}
      &\multicolumn{1}{c}{overall} & \multicolumn{1}{c}{things} & \multicolumn{1}{c}{stuff} & \multicolumn{1}{c}{singulars} & \multicolumn{1}{c}{plurals} \\ \hline\hline
     \checkmark & & & 66.0 & 62.7 & 70.5 & 66.9 & 57.6 \\
     \checkmark & \checkmark & & 66.7 & 63.7 & 70.9 & 67.5 & 59.9 \\
     \checkmark & \checkmark & \checkmark & \textbf{67.1} & \textbf{64.3} & \textbf{71.0}  & \textbf{67.9} & \textbf{60.0} \\
     \hline
    \end{tabular}}
    }
    \caption{Ablations of components in MLMA module.}
    \label{tab:ablation:mlma_ops}
\end{table}

\textbf{MLMA Component Analysis.}
We also conducted ablation studies to assess the impact of different operations within the MLMA module on segmentation performance, with results shown in Table~\ref{tab:ablation:mlma_ops}.
Building on the application of deformable attention to multi-level image features, introducing multi-level phrase features and applying bi-attention between image and phrase features lead to performance improvements.
This indicates that incorporating textual modal clues into multi-scale information of images is beneficial for predicting segmentation masks.
Furthermore, applying self-attention layers to phrase features for feature enhancement, thereby iteratively updating the phrase features within the bi-attention, results in further improvements in segmentation performance.

\subsection{Qualitative Results}
As shown in Figure~\ref{fig:vis:adapter_attn}, we visualize the cross-attention maps in different layers of our EIPA.
We conduct $\mathrm{Softmax}$ on the phrase dimension of the cross-attention map and select the $\mathrm{arg~max}$ phrase label for each pixel, which approximately shows which phrase is the most matched to each pixel.
Take the 2-nd row as an example, we can observe that attention maps in the shallow layers (\textit{e.g.}, $L=3$ or $5$) of EIPA have low resolutions and distribute relatively scattered on the referred objects.
While in the 12-th layer of EIPA, the resolution is recovered and pixels are matched with the correct phrases, which indicates cross-attentions in our EIPA can capture precise correlations between image and text.

Figure~\ref{fig:vis:pred_vs_gt} presents the qualitative results of our proposed method, where different colored segmentation mask regions correspond to phrases of matching colors.
Our method is capable of generating high-quality segmentation masks based on dense textual descriptions.
Areas where the predictions of our method differ from the ground-truth annotations are highlighted with white dashed boxes
For instance, in the first row, the ground-truth annotation misses distant pedestrians and incorrectly labels the main subject's arm, whereas our method accurately segments these pedestrians.
In these examples, our method is capable of predicting fine segmentation masks and correctly associating them with the correct phrases, demonstrating the effectiveness of dynamically prompting the large text-to-image Diffusion models.

\section{Conclusion}
We study the PNG task where previous discriminative methods achieve only weak or coarse alignment via panoptic segmentation pretraining or adapting the CLIP model.
Recently, many studies have demonstrated the success of text-to-image Diffusion models in attaining fine-grained image-text alignment.
However, static prompting of Diffusion models using fixed phrase features still suffers from a large task gap and insufficient vision-language interaction when adapted to the PNG task.
Therefore, we propose an EIPA bypass to dynamically update phrase prompts with image features and inject the multimodal cues back, leading to more sufficient fine-grained image-text alignment.
We also develop an MLMA module to refine segmentation quality via reciprocal fusion of multi-level features.
Our method achieves state-of-the-art performance on the PNG benchmark.

\begin{acks}
This research was supported in part by National Science and Technology Major Project (2022ZD0115502), National Natural Science Foundation of China (No. 62122010, U23B2010, 62132001), Zhejiang Provincial Natural Science Foundation of China (Grant No. LDT23F02022F02), National Key Research and Development Program of China (No. 2021YFB1714300), Beijing Natural Science Foundation (No. L231011), Beihang World TOP University Cooperation Program, and Meituan.
\end{acks}

\bibliographystyle{ACM-Reference-Format}
\bibliography{sample-base}

% \appendix









\end{document}